# DeepQC: A Deep Learning System for Automatic Quality Control of In-situ Soil Moisture Sensor Time Series Data

Lahari Bandaru[1]*, Bharat C Irigireddy[2], Brian Davis[3]*

[1] Poolsville High School, Poolesville, MD;
[2] A. James Clark School of Engineering, University of Maryland, MD
[3] Precission Sustainable Agriculture, Beltsville, MD

E-mail: laharibandaru@gmail.com, bwdavis3@ncsu.edu

**Abstract**

Amidst changing climate, real-time soil moisture monitoring is vital for the development of in-season decision support tools to help farmers manage weather-related risks in agriculture. Precision Sustainable Agriculture (PSA) recently established a real-time soil moisture monitoring network across the central, Midwest, and eastern U.S., but continuous field-scale sensor observations often come with data gaps and anomalies. To maintain the data quality and continuity needed for developing decision tools, a quality control system is necessary.

The International Soil Moisture Network (ISMN) introduced the Flagit module for anomaly detection in soil moisture time series observations. However, under certain conditions, Flagit's threshold and spectral based quality control approaches may underperform in identifying anomalies. Recently, deep learning methods have been successfully applied to detect time series anomalies in time series data in various disciplines. However, their use in agriculture for anomaly detection in time series datasets has not been yet investigated. This study focuses on developing a Bi-directional Long Short-Term Memory (LSTM) model, referred to as DeepQC, to identify anomalies in soil moisture time series data. Manual flagged PSA observations were used for training, validation, and testing the model, following an 80:10:10 split. The study then compared the DeepQC and Flagit-based estimates to assess their relative performance.

Flagit correctly flagged 95.5% of the correct observations and 50.3% of the anomaly observations, indicating its limitations in identifying anomalies, particularly at sites consists of more than 30% anomalies. On the other hand, the DeepQC correctly flagged 99.7% of the correct observations and 95.6% of the anomalies in significantly less time, demonstrating its superiority over Flagit appraoch. Importantly, the performance of the DeepQC remained consistent regardless of the number of anomalies in site observations. Given the promising results obtained with the DeepQC, future studies will focus on implementing this model on national and global soil moisture networks.

**1. Introduction**

The changing climate has led to more frequent extreme events, including droughts and heavy rainfall. These extreme events present significant challenges for farmers in making timely decisions, ultimately jeopardizing both agricultural production and the well-being of farmers globally (Raj et al., 2022). Given the anticipated unusual growing conditions due to climate change, it is crucial to develop operational decision support systems that can help farmers make informed decisions to mitigate weather risks and improve agricultural production. Real-time soil moisture monitoring plays a critical role in supporting farm or field-scale agricultural decisions, as it directly affects crop growth, yield, and resource management (Cosh et al., 2021). Timely information about soil moisture levels in the root zone can assist in irrigation scheduling. Additionally, it aids in crop selection and optimization of in-season management practices. Knowledge of soil moisture levels at the beginning of the growing season helps determine the best crops to cultivate based on seasonal conditions. Similarly, in-season soil moisture information assists in forecasting crop nutrient requirements and determining precise fertilizer application rates. This information can also aid in the early prediction of pest and disease infestations, allowing for timely intervention when necessary. Furthermore, field-scale soil moisture observations are required as reference data for the calibration and validation of satellite-based soil moisture products, such as Soil Moisture Active Passive (SMAP), Soil Moisture and Ocean Salinity (SMOS), and Advanced Microwave Scanning Radiometer 2 (AMSR2) (Al-Yaari et al., 2019; Beck et al., 2021).

Soil moisture sensors are essential tools for the automated monitoring of field-scale soil moisture, providing continuous real-time data required for developing decision support systems. A wide range of measurement techniques is available for collecting continuous observations in agricultural fields. Commonly used techniques include Time-Domain Reflectometer (TDR), electrical resistance sensors, neutron probes, tensiometers, capacitance sensors, and cosmic ray neutron sensors (Mittelbach et al., 2012; Lekshmi et al., 2014; Babaeian et al., 2019; Rasheed et al., 2022). Even though each type has its advantages and suitable use cases, observations often exhibit data gaps and uncertainties, regardless of the sensor type (Xia et al., 2015). Some sources causing data gaps and anomalies include poor sensor calibration, incorrect installation, adverse environmental conditions (e.g., excessive moisture or extreme temperatures), poor maintenance, inadequate power supply, and network congestion (Dorigo et al., 2013; Xaver, 2015; Liao et al., 2019). The magnitude of gaps and anomalies depends on various factors such as vegetation type, prevailing weather conditions, and soil characteristics, as well as their interactions (Xaver et al., 2015). Since sensor data quality significantly impacts data interpretation, modeling, and subsequent decision-making, it is critical to have the ability to identify data gaps and anomalies in observations to ensure data continuity and high quality (Campbell et al., 2013).

Visual inspection is the most effective way to ensure the quality of observational data (Dorigo et al., 2013). However, the large volume of data produced by sensors on a daily basis often makes manual inspection impractical. Therefore, a feasible solution would be to develop automatic Quality Control (QC) approaches capable of automatically detecting anomalies and data gaps and flagging them. Given the significance of data quality, various automatic QC approaches have been developed. The most common QC approaches use threshold values derived based on the typical behavior of soil moisture data. These threshold values can be constant, time-variable, or determined using statistical characteristics such as mean and standard deviation (Journée and Bertrand, 2011; Dorigo et al., 2013; Liao et al., 2019). For instance, in the North American Soil Moisture Database (NASMD) (Quirring et al., 2016), the QC procedure identifies spurious soil moisture observations that fall outside the geophysical range of 0 to 0.6 (m3/m3), remain constant for 10 consecutive days, or exceed three standard deviations (SD) during a 30-day period (Liao et al., 2019). Spectrum-based methods identify sudden, unusual, erroneous changes such as spikes, breaks, and constant values within the soil moisture time series (Dorigo et al., 2013; Xaver et al., 2015).

Even though existing methods have demonstrated reasonable performance on regional to global scales, they often fail to identify erroneous data under specific circumstances. For instance, Dorigo et al. (2013) reported low flagging accuracy with threshold-based methods for detecting erroneous events under certain conditions. Using precipitation data from the Global Land Data Assimilation System (GLDAS) for flagging observations resulted in an accuracy of only 52%. Similarly, when employing global soil data to assess geophysical consistency, the accuracy of flagging dropped to 37%. The poor performance of these geophysical constraints was attributed to uncertainties in global soil and precipitation datasets. Relying on coarse-resolution ancillary inputs can lead to the poor performance of threshold methods for flagging. Furthermore, Dorigo et al. (2013) found that their spectrum-based method failed to identify spikes if the spikes lasted longer than one time step or were surrounded by missing values. Additionally, it incorrectly flagged numerous observations with data breaks. Given these challenges in existing methods, there is a need for alternative approaches capable of identifying data gaps and anomalies under various circumstances with minimal reliance on ancillary data, ensuring robust quality control of soil moisture observations.

Deep Learning Neural Networks (DNN) have the ability to capture complex patterns and irregularities in time series data, enabling them to identify subtle anomalies (Choi et al., 2021). Recently, Long Short-Term Memory (LSTM), a recurrent neural network model, has been successfully applied to detect anomalies in various time series datasets, including physiological signal data (e.g., electrocardiogram or ECG), communication network data, and sensor data from industrial factories (Buda et al., 2018; Nguyen et al., 2018; Zhang and Zou, 2018; Zhang et al., 2019; Ji et al., 2021; Saadallah et al., 2022; Li and Jung, 2023). The LSTM architecture is designed to capture long-term dependencies in sequential data. Its ability to learn complex patterns and evolving dependencies over time in time series data, while maintaining them over extended sequences, makes it well-suited for recognizing subtle anomalies that may span multiple time steps (Gopali and Namin, 2022). Despite LSTM's promise in identifying anomalies in time series data, to the best of our knowledge, it has not been previously investigated for quality control of sensor time series data in agriculture.

Considering the challenges in existing quality control (QC) methods and the promising potential of LSTM for anomaly detection, the goal of this study is to develop an LSTM-based deep learning framework, hereafter referred to as DeepQC, for identifying anomalies in in-situ soil moisture time series data. We will evaluate its performance by comparing it to the anomalies detected by the Flagit module (Dorigo et al., 2013). To develop and test the LSTM model, we used manually flagged soil moisture time series data from the Precision Sustainable Agriculture Soil Moisture Network. The specific objective of this study is to use only soil moisture time series data for LSTM model development, without including ancillary variables such as precipitation and air temperature, which are typically used in quality control methods.

Carolina and United States Department of Agriculture Agricultural Research Service. PSA team integrates sustainable agriculture and precision techniques to develop decision tools for farmers to improve production sustainably. The PSA established a soil moisture network in 2017, deploying soil moisture sensor suits over 663 farmers fields across the Central and Eastern U.S, of which. The goal of this network is to observe the soil moisture patterns in various soil and crop types to help with crop selection and in-season management decisions of the farmers. Each sensor suite consists of four Acclima True TDR-310H soil water content sensors, placed in four different depths (i.e., 0-5 cm (surface depth), 0-30 cm, 30-60 cm and 60-100 cm) (Figure 1a). All sensor readings are communicated wirelessly and then stored on a central web application programming interface (API).

For this study, 98 sites containing manual flags were selected (Figure 1b). Each site has four sub-plots, two treatments (crop and bare fields) and two replications. The soil moisture and soil temperature data are available from 2017-present. The crops cultivated in these fields include corn, soybeans and cotton and most of the fields are under rainfed agriculture. The elevation ranges from 5-643 m, and average annual precipitation (averaged over 2017-2021) ranges from 788-1758 mm. Site characteristics for each field is provided in Appendix 1.

## 3. Methodology

The methodology involved in developing DeepQC based

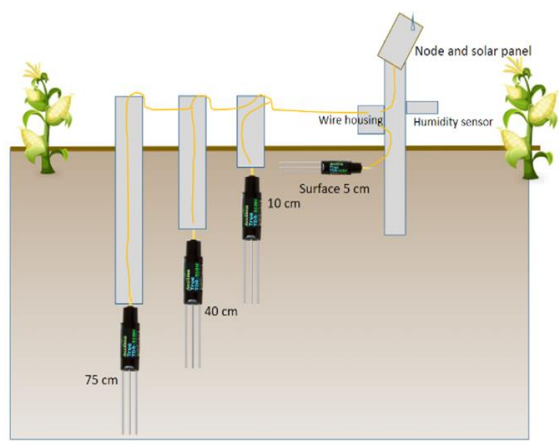
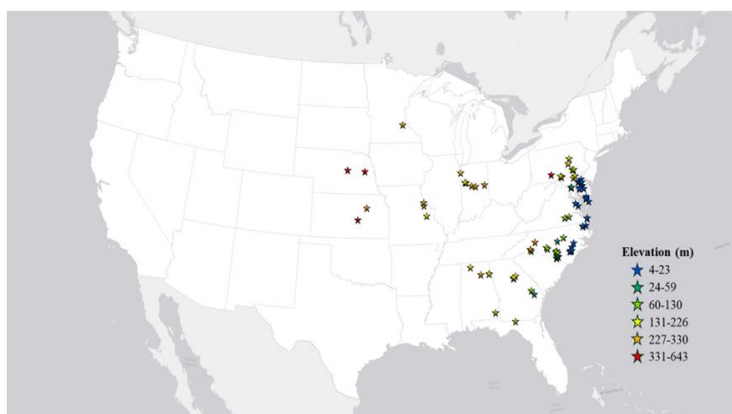

Fig. 1. Schematic of sensor suite consisting of four Acclima True TDR-310H soil water content sensors, placed in four different depths (i.e., 0-5 cm (surface depth), 0-30 cm, 30-60 cm and 60-100 cm) (**A**); PSA field sites used in this study where manually flagged soil moisture data is available (**B**).

## 2. Precision Sustainable Agriculture Soil Moisture Network

Precision Sustainable Agriculture (PSA) is a network of researchers from different agencies and universities, including Cornell University, Purdue University, North on LSTM and its performance evaluation is illustrated in Figure 2. Firstly, the ISMN Flagit module was used to flag the correct and anomaly observations in the PSA data. Further, the PSA data from 98 field sites were divided into training, validating, and testing datasets with an 80:10:10 split. The DeepQC model, featuring bidirectional LSTM, was trained on soil moisture sequences of length 96 that represent a full day's data at 15-minute intervals to ensure a detailed temporal resolution. The model's performance was monitored by calculating the loss function on the validation set after each epoch. For the testing phase, the DeepQC model's accuracy was assessed using a binary confusion matrix, contrasting its predictions with manual anomaly flags present in the test data. To address the issue of imbalanced data, days containing at least one anomaly were given more weight during the training process. This approach enhances the model's sensitivity to detect irregularities. Furthermore, the results of the LSTM model and the ISMN Flagit module were compared using PSA manually flagged observations.

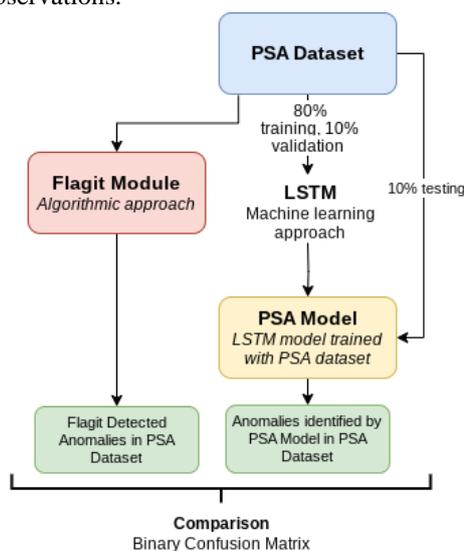

Fig. 2. Schematic illustrating steps for DeepQC based on Long Short-Term Memory (LSTM) development and testing its performance against the Flagit module.

### 3.1 Potential Erroneous Events within PSA Soil Moisture Readings

Soil moisture time series exhibits few typical characteristics that are common for the most records. For example, a typical soil wetting event immediately after precipitation causes a sharp rise in soil moisture which gradually decreases during consecutive drying until it reaches saturation level (Figure 3a).

Similarly, during freezing events, lower soil moisture is recorded due to significantly lower dielectric conductivity of ice compared to liquid water (Dorigo et al., 2013). A variety of anomalies may occur in soil moisture time series data due to various factors such as data logger issues and sensor failure due extreme weather conditions. We noticed spikes and breaks more often in PSA soil moisture data. A spike is an anomaly of typically a single or couple of observation values which vary significantly from the values before and after (Figure 3b). A break is a sharp rise or drop of the observation compared to measurement before, and the time series before and after break would appear continuous without the anomaly event (Figure 3c). Spikes and breaks in PSA data mostly likely result from interruption in data logger communication with a reduced power supply or sudden change in weather conditions.

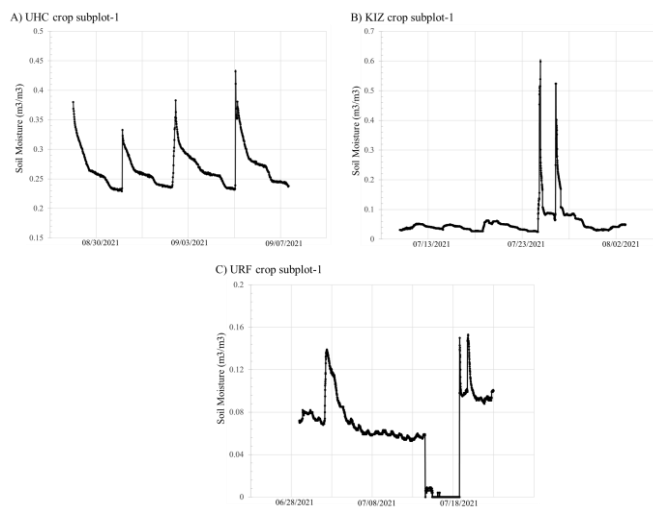

Fig. 3. Examples of surface (5cm) soil moisture time series for three PSA crop subplots containing a) typical wetting and drying event b) spikes and c) breaks.

### 3.2 Description of Flagit module

The Flagit is a Python module developed based on the soil moisture quality control system created by Dorigo et al. (2013). Flagit module is available in the GitHub repository (Daberer, 2022). This QC system employs a combination of threshold and spectrum-based methods to identify various anomalies in soil moisture data. The threshold methods assess the geophysical dynamic range and data consistency using various thresholds listed in Table 1. Additionally, spectrum-based techniques are utilized to detect spikes, breaks, and constant values. For time series analysis, a Savitzky-Golay filter (Savitzkey and Golay, 1964) is applied to smooth soil moisture data, and calculates first and second derivatives. By examining the characteristic patterns of anomalies within the first and second derivatives of the time series, different conditions and empirical equations were developed to detect spikes, breaks, and constant values. A more detailed description of the QC system and its equations can be found in Dorigo et al. (2013).

### 3.3 Application of Flagit module to PSA data

The manually flagged PSA data was used to detect anomalies using the Flagit module. The PSA data includes soil temperature measurements for each depth, but it lacks records for precipitation and air temperature, which are needed to assess the geophysical dynamic range and data consistency required to identify specific anomalies.

Table 1. Flagging criteria in the Flagit module based on thresholds to detect anomalies.

| Flag | Criteria |
|---|---|
| C01 | Soil moisture value is less than 0 m³/m³ |
| C02 | Soil moisture value is greater than 0.60 m³/m³ |
| C03 | Soil moisture value is greater than the saturation point, which is calculated using ancillary sand, clay and organic content |
| D01/D03 | Soil temperature < 0°C at corresponding depth. D01 is labelled if in-situ soil temperature is considered. If GLDAS data is used, D03 is labelled. |
| D02 | In-situ air temperature < 0°C |
| D04/D05 | Rise in soil moisture values without a corresponding rise in ancillary in situ precipitation within the previous 24 hours. D04 is labelled if in-situ soil moisture is considered. If GLDAS data is used, D05 is labelled. |

To obtain precipitation and air temperature data, we used the North American Land Data Assimilation System (NLDAS) hourly meteorological data, which has 1/8-degree, approximately 12 km resolution (Cosgrove et al., 2003). The PSA data consists of observations at 15 min intervals so each 15-minute PSA observation falling within a 1-hour time range of an NLDAS observation is assigned with the corresponding NLDAS observation.

Given that each agricultural site has a unique soil moisture sensor allocated to each depth, the soil moisture dataset is initially categorized by site code and center depth, resulting in sequential data organized by soil moisture sensor. This data is passed into the Flagit module, and an additional column labeled 'qflag' is added to the datafile to indicate the flags associated with each data entry. The program then applies the Flagit module and stores the corresponding flags.

### 3.4 Description of Long Short-Term Memory (LSTM)

Long Short-Term Memory (LSTM) is a type of recurrent neural network (RNN) designed to handle sequential data. RNN models take the output from the previous step as input for the current step. Traditional RNN models are thus capable of making accurate predictions based on previous information, but only if that information is from recent steps. The greater the distance between the current step and a previous step, the less relevant that previous step becomes to the current output. As such, traditional RNN models are incapable of handling long term dependencies.

LSTM models solve this using a memory cell, or cell state, which stores information persistently across iterations. LSTM is thus suitable for this research project, which attempts to identify anomalies depending on dependencies in long term soil moisture patterns.

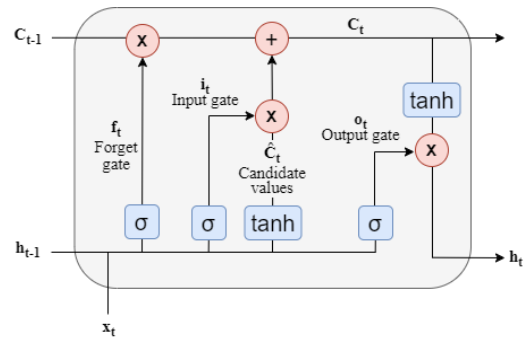

Fig.4. Schematic illustrating structure the Long Short-Term Memory (LSTM).

### 3.4.1 LSTM model structure

The LSTM model structure consists of several neural network "gates" and a memory cell, also known as the cell state. The memory cell stores information across iterations, and gates manage what information is stored in the memory cell. There are three gate types: the forget gate which determines what information is removed from the memory cell, and the output gate which determines what values the cell state outputs.

1. Forget Gate: The forget gate determines what information is no longer useful, to be purged from the memory cell. The forget gate takes two inputs, both the current time input and the model outputs from the previous iteration. Weight matrices are applied to these inputs, along with a bias value. The output is passed through a sigmoid activation function to offer an output between 0 and 1, signifying what percentage of that memory should continue to be stored: 0 representing extraneous information that should be completely removed from the memory cell, and 1 representing significant information that should be maintained by the memory cell.

$$f_t = \sigma(W_{xf} \cdot x_t + b_{xf} + W_{hf} \cdot h_{t-1} + b_{hf})$$

*Where $f_t$ is the output from the forget gate. $x_t$ is the input for the current iteration, and $h_{t-1}$ is the output from the previous iteration. $W_{xf}$ is the forget gate's weight for the current input. $W_{hf}$ is the forget gate's weight for the output from the previous iteration. $\sigma$ represents the activation function. $b_f$ represents the forget gate's associated bias.*

2. Input Gate: Information is classified as useful and stored in the cell state through the input gate. The input gate, like the forget gate, accepts the output from the previous step, the input for the current step, applies a weight matrix and a bias, and passes the result through a sigmoid activation function to retrieve a boolean output.

$$i_t = \sigma(W_{xi} \cdot x_t + b_{xi} + W_{hi} \cdot h_{t-1} + b_{hi})$$

*Where $i_t$ represents the extent to which the cell state must be updated from the current iteration.*

Next, a tanh layer is applied to the same input, creating a vector of all values h$_{t-1}$ and x$_t$ which may potentially be added to the memory cell.

$$\hat{C}_t = tanh(W_{xC} \cdot x_t + b_{xC} + W_{hC} \cdot h_{t-1} + b_{hC})$$

*Where $\hat{C}_t$ represents the new candidate values.*

$f_t$ is applied to the previous memory cell state $\hat{C}_{t-1}$ to remove all extraneous memory, while i$_t$ is applied to $\hat{C}_t$ to retrieve memory to be stored in the cell state. The results are summed to produce the updated cell state.

$$C_t = f_t * C_{t-1} + i_t * \hat{C}_t$$

3. Output Gate: The output gate determines the model output. First, h_(t-1) and x_t are passed through a sigmoid layer to determine what parts of the memory cell will be outputted.

$$o_t = \sigma(W_{xo} \cdot x_t + b_{xo} + W_{ho} \cdot h_{t-1} + b_{ho})$$

*Where $o_t$ represents the extent to which the cell state will be outputted.*

The memory cell is passed through a tanh activation function to fit the values between -1 and 1, before o$_t$ is applied to the current cell state to retrieve the output value.

$$h_t = o_t * tanh(C_t)$$

*Where $h_t$ represents the model output for the current step.*

## 3.5 Description of Long Short-Term Memory (LSTM)

The DeepQC was designed for anomaly detection in soil moisture sensor readings, focusing on identifying long-term dependencies that could indicate anomalies in the measurements. The model takes a sequence of soil moisture readings as input, along with additional context features: center depth and day of the year.

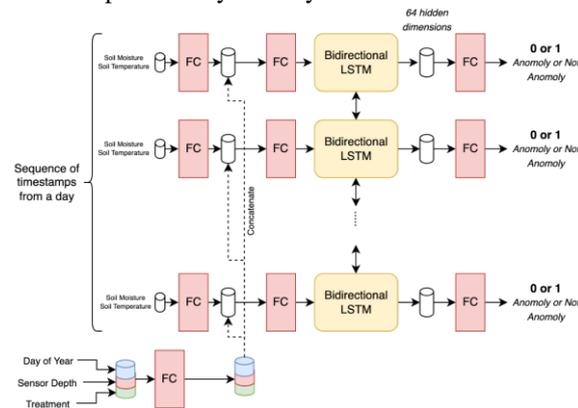

Fig. 5. Architecture of the DeepQC Long Short-Term Memory (LSTM) model.

Two initial linear layers expand the features into a unified 32-dimensional feature space. These features were then processed by a bidirectional LSTM layer with a hidden dimension of 64 to capture temporal dependencies (Figure 5). The output of the LSTM was funneled through a final linear layer and a sigmoid activation to yield a probability score between 0 and 1, indicating the likelihood of each observation being an anomaly. The model was trained for 400 epochs, and the loss function was estimated for each epoch using the validation dataset.

## 3.6 Training and validation of DeepQC

### 3.6.1 Binary Cross Entropy Loss

Binary cross entropy loss is used to determine the performance of a binary classification model. Cross entropy quantifies the difference between two probability distributions given random events.

$$-(ylog(p) + (1-y)log(1-p))$$

*Where log is the natural log, y is the binary indicator (0 or 1), and p is the probability of the selected binary indicator.*

### 3.6.2 Adam Optimization Algorithm

The optimization algorithm is tasked with updating network weights to minimize the loss function. The Adam optimization algorithm extends the rudimentary stochastic gradient descent optimization algorithm by utilizing both exponential moving average of gradients and square gradients. The parameters β_1 and β_2 are used to dictate the decay rate of these moving averages. Adam is known for being very effective in handling sparse gradients, and for having a very straightforward approach. As such, the Adam optimization algorithm is The Adam optimization algorithm is typically considered the default optimization method for deep learning applications as a result of its simplicity and flexibility for various machine learning model structures.

## 3.7 Comparison between Flagit and DeepQC estimates

### 3.7.1 Binary confusion matrix

Accuracy analysis was conducted using binary confusion matrices, which plot a model's true positive, false positive, true negative, and false negative count. A binary confusion matrix was generated using the manual labels for the PSA soil moisture data and the flags predicted by the Flagit module.

Table 2. Structure of confusion matrix.

| *C* | *0* | *1* |
|---|---|---|
| *0* | $C_{0,0}$: True negatives | $C_{0,1}$: False positives |
| *1* | $C_{1,0}$: False negatives | $C_{1,1}$: True positives |

## 4. Results and Discussion

### 4.1 Flagging Performance of Flagit module

The total number of manually flagged observations is 892,447, of which 852,127 (95.5%) are correct observations, and the remaining 40,320 (4.7%) are anomalies. Most of the anomalies are characterized by values outside of geophysical ranges, spikes and breaks. Flagging performance of the Flagit module was assessed by comparing it with manual flags and using a classical error matrix (Congalton and Green, 2009). The results were summarized in Table 3. The results showed that Flagit correctly flagged most of the correct observations with accuracy more than 95%, and only 4.2% of them were flagged incorrectly. However, the detection accuracy of anomalies was considerably low with only 50.3% of the total anomalies being correctly flagged.

Table 3. Summary of flagging performance of Flagit module based on manual flags.

| Flagging Results | Correctly Flagged | Incorrectly Flagged | Total |
|---|---|---|---|
| Correct | 816,764 | 35,363 | 852,127 |
| Correct (%) | 95.85 | 4.15 | 100.00 |
| Anomalies | 20,272 | 20,047 | 40,319 |
| Anomalies (%) | 50.28 | 49.72 | 100.00 |
| Total | 837,036 | 55,410 | 892,447 |
| Total (%) | 93.79 | 6.21 | 100.00 |

Our study revealed a substantial decrease in the detection accuracy of anomalies for sites with more than 30% anomalies in the observations. Figure 6 illustrates two examples showing the variation in detection accuracy with a distant number of anomalies present in the observations. In the first example (Figure 6a & b), anomalies constituted 43.6% of the data, and Flagit failed to detect 98% of the anomalies. In the second example (Figure 6c & d), anomalies comprised less than 1%, and all of them were successfully identified. Both examples exhibited spikes in the observations. However, in the first example with higher anomaly levels, Flagit did not flag the spikes, whereas it correctly detected the spikes in the second example. Additionally, in the first example, breaks were not detected. Flagit employs spectral methods for identifying spikes and breaks (Dorigo et al., 2013). These methods utilize the Savitzky-Golay filter to smooth time series data and calculate their first and second derivatives. Peaks and their magnitudes in the derivative curves are used to identify spikes and breaks. The high number of anomalies negatively impacts the performance of the Savitzky-Golay filter in smoothing the data and identifying peaks and their magnitudes, resulting in poorer performance of the spectral methods for identifying spikes and breaks. Previous studies (Dorigo et al., 2013 and Xia et al., 2015) have also emphasized the limitations of spectral methods in the presence of inconsistent time series data.

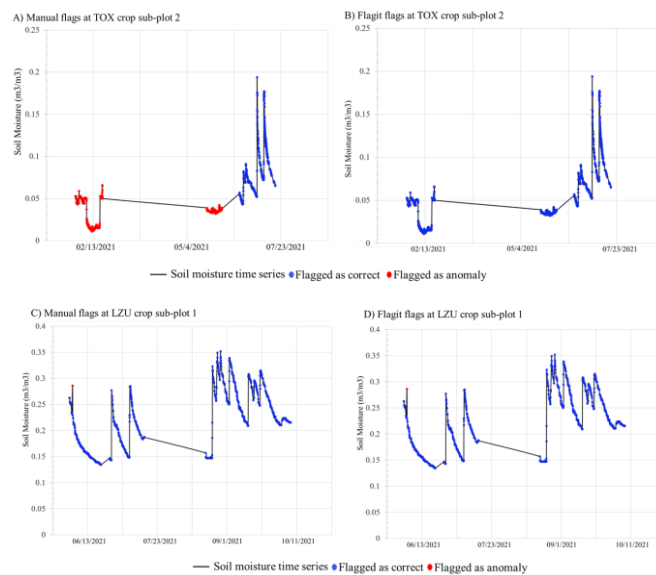

Fig.6. Comparison of manually and Flagit flagged soil moisture observations for two distant sites with varied number of anomalies. TOX site (top) has anomalies for 43.6% of total observations whereas LZU site (bottom) has anomalies for less than 1% observations. a) manual flags for TOX crop subplot-1; b) Flagit flags for TOX crop subplot-1; c) manual flags for LZU crop subplot-1; d) Flagit flags for LZU crop subplot-1.

### 4.2 Performance of DeepQC model

Two-level performance assessment was conducted: 1) validation during model development and 2) validation after implementation. The training and validation accuracies and losses over the training period are shown in Figure 7. It can be observed that as the training progressed with epochs, the training loss was reduced, indicating that the DeepQC classifier learned to fit the training data accurately. A similar curve pattern for validation loss and accuracy implies that the DeepQC can generalize its performance when implemented in different sites by matching the pattern in the training curves. The best validation performance was observed to be at epoch 620, after which we noticed a slight overfitting. This could be attributed to low fraction of anomalies.

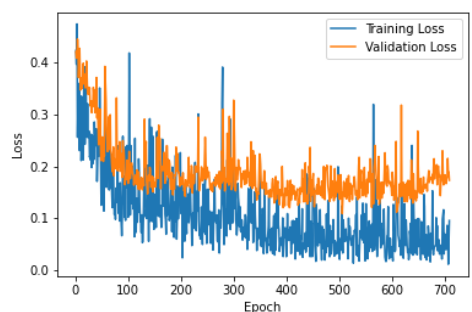

Fig.7. Training and validation losses of DeepQC.

We used this best model to evaluate the performance using the test data. The results showed that the test accuracy of the final model is 95.6%.

Table 4. Summary of flagging performance of DeepQC model represented by confusion matrix.

| DeepQC Results | | | |
|---|---|---|---|
| | Correctly Marked | Incorrectly Marked | Total |
| Correct | 35034 | 3966 | 39000 |
| Correct (%) | 89.83 | 10.17 | 100 |
| Anomalies | 49983 | 261 | 50244 |
| Anomalies (%) | 99.48 | 0.52 | 100 |
| Total | 85017 | 8380 | 89244 |
| Total (%) | 95.61 | 9.39 | 100 |

In contrast to Flagit module, DeepQC performed well irrespective of number of anomalies present in the soil moisture measurements. In both examples (Figure 8a&b), DeepQC was able to flag both correct and anomaly observations correctly. There were few anomaly observations flagged as correct at TOX sub-plot which could be result of slight overfitting (Figure 8a).

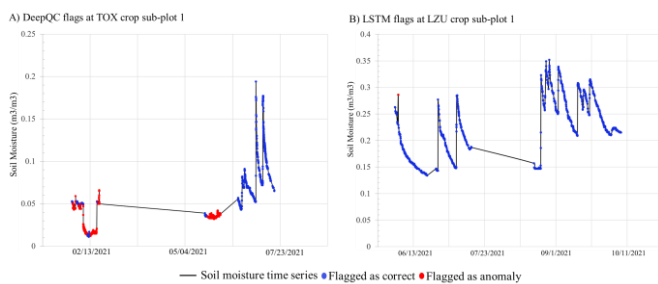

Fig.8. Flagging performance of DeepQC model at TOX site (a) and LZU site (b).

Furthermore, we evaluated time efficiency of DeepQC model compared to Flagit model. The Flagit model took approximately 12 min for 150,000 selected observations while DeepQC model took approximately 30 sec indicating significantly greater time efficiency of DeepQC model.

## 5. Conclusions

Current threshold and spectral-based quality control methods have limitations in identifying anomalies when many anomalies occur in soil moisture measurements. This limitation was evident in our study with the Flagit module, which identified only 50.3% of the total anomalies in PSA soil moisture manually flagged observations. Furthermore, these methods require ancillary variables such as soil and climate data, which are often characterized by uncertainties affecting anomaly detection accuracy. The DeepQC developed based on Bi-directional Long Short-Term Memory (LSTM) model using the Precision Sustainable Agriculture manually flagged soil moisture dataset without using any ancillary variables, outperformed the Flagit module with an anomaly detection accuracy of 96.5% and higher time efficiency. This demonstrated that it is more suitable for identifying anomalies in soil moisture datasets.

However, the model does have a few shortcomings and areas for improvement. The most notable weakness is the low percentage of anomalies in the dataset. While this is characteristic of realistic datasets, it hinders the model's learning capacity to some extent. Learning capacity can be greatly increased with the use of synthetic data introducing different possible anomaly types such as spikes, breaks and constant values to provide a larger anomaly dataset for the model to learn from.

Another potential experiment is to investigate time series transformers as an alternative to the LSTM model. Transformers-based anomaly detection models, like LSTM models, are used for sequential data analysis. However, unlike LSTM models and other recurrent neural network models, transformers avoid the recursive analysis of previous inputs, enabling parallel computation of data input. As such, transformers are typically far more efficient and effective at determining long-term dependencies.

Additionally, there are numerous public soil moisture datasets available for further refining the model's performance at independent locations worldwide. These include the North American Soil Moisture database, which includes 800 monitoring stations in the United States, Canada, and Mexico, as well as the International Soil Moisture Network, consisting of data from more than 71 networks and 2,842 stations from across the globe, spanning from 1952 to the present.

## Acknowledgements

The authors thank Mr. Zachary S Kingman, Poolsville High School; Dr. Koutilya PVNR, University of Maryland; and Mr. Mikah Pinegar, Precision Sustainable Agriculture for their valuable suggestions and support.